\documentclass[letterpaper, 10 pt, conference]{ieeeconf}   %

\IEEEoverridecommandlockouts                              %

\usepackage{graphics} %
\usepackage{epsfig} %
\usepackage{cuted}
\usepackage{caption}
\usepackage{amsmath} %
\usepackage{import}
\usepackage{color}
\usepackage{bm}
\usepackage{hhline}
\usepackage{multirow}
\usepackage[hidelinks]{hyperref}
\usepackage{float}
\usepackage[table]{xcolor}
\usepackage{flushend}
\usepackage{multicol}

\DeclareMathOperator*{\argmin}{arg\,min}

\title{ \LARGE \bf
Single-View and Multi-View Depth Fusion
}

\author{Jos\'e M. F\'acil$^{1}$, Alejo Concha$^{1}$, Luis Montesano$^{1,2}$ and Javier Civera$^{1}$
\thanks{We gratefully acknowledge the support of NVIDIA Corporation for the donation of a Titan X GPU, the Spanish
government (projects DPI2012-32168 and DPI2015-67275),
the Aragon regional government (Grupo DGA T04-FSE) and ´
the University of Zaragoza (JIUZ-2015-TEC-03).}%
\thanks{$^{1}$The authors are with the I3A, University of Zaragoza, Spain
        {\tt\small \{jmfacil, montesano, jcivera\}@unizar.es},  %
        {\tt\small aconchabelenguer@gmail.com}}
\thanks{$^{2}$Luis Montesano is also with Bit\&Brain Technologies SL.}
}

\begin{document}
\twocolumn[{\centering This paper has been accepted for publication in \textit{IEEE Robotics and Automation Letters}.\\}
\vspace{12pt}
\begin{center}
DOI:10.1109/LRA.2017.2715400\\
IEEE Xplore: \href{http://ieeexplore.ieee.org/document/7949041/}{{\tt http://ieeexplore.ieee.org/document/7949041/}}
\end{center}
\vspace{24pt}
\textcopyright \ 2017 IEEE. Personal use of this material is permitted. Permission from IEEE must be obtained for all other uses, in any current or future media, including reprinting /republishing this material for advertising or promotional purposes, creating new collective works, for resale or redistribution to servers or lists, or reuse of any copyrighted component of this work in other works.]

\maketitle
\thispagestyle{empty}
\pagestyle{empty}
\begin{abstract}

Dense and accurate 3D mapping from a monocular sequence is a key technology for several applications and still an open research area. This paper leverages recent results on single-view CNN-based depth estimation and fuses them with multi-view depth estimation. Both approaches present complementary strengths. Multi-view depth is highly accurate but only in high-texture areas and high-parallax cases. Single-view depth captures the local structure of mid-level regions, including texture-less areas, but the estimated depth lacks global coherence.
The single and multi-view fusion we propose is challenging in several aspects. First, both depths are related by a 
deformation that depends on the image content. Second, the selection of multi-view points of high accuracy might be difficult for low-parallax configurations. We present contributions for both problems. Our results in the public datasets of NYUv2 and TUM shows that our algorithm outperforms the individual single and multi-view approaches. A video showing the key aspects of  mapping in our Single and Multi-view depth proposal is available at \href{https://youtu.be/ipc5HukTb4k}{{\tt https://youtu.be/ipc5HukTb4k}}.

\end{abstract}


\section{INTRODUCTION}
Estimating an online, accurate and dense 3D scene reconstruction from a general monocular sequence is one of the fundamental research problems in computer vision. The problem has nowadays a high relevance, as it is a key technology in several emerging application markets (augmented and virtual reality, autonomous cars and robotics in general).
\begin{figure}
\centering
\includegraphics[width=1\linewidth]{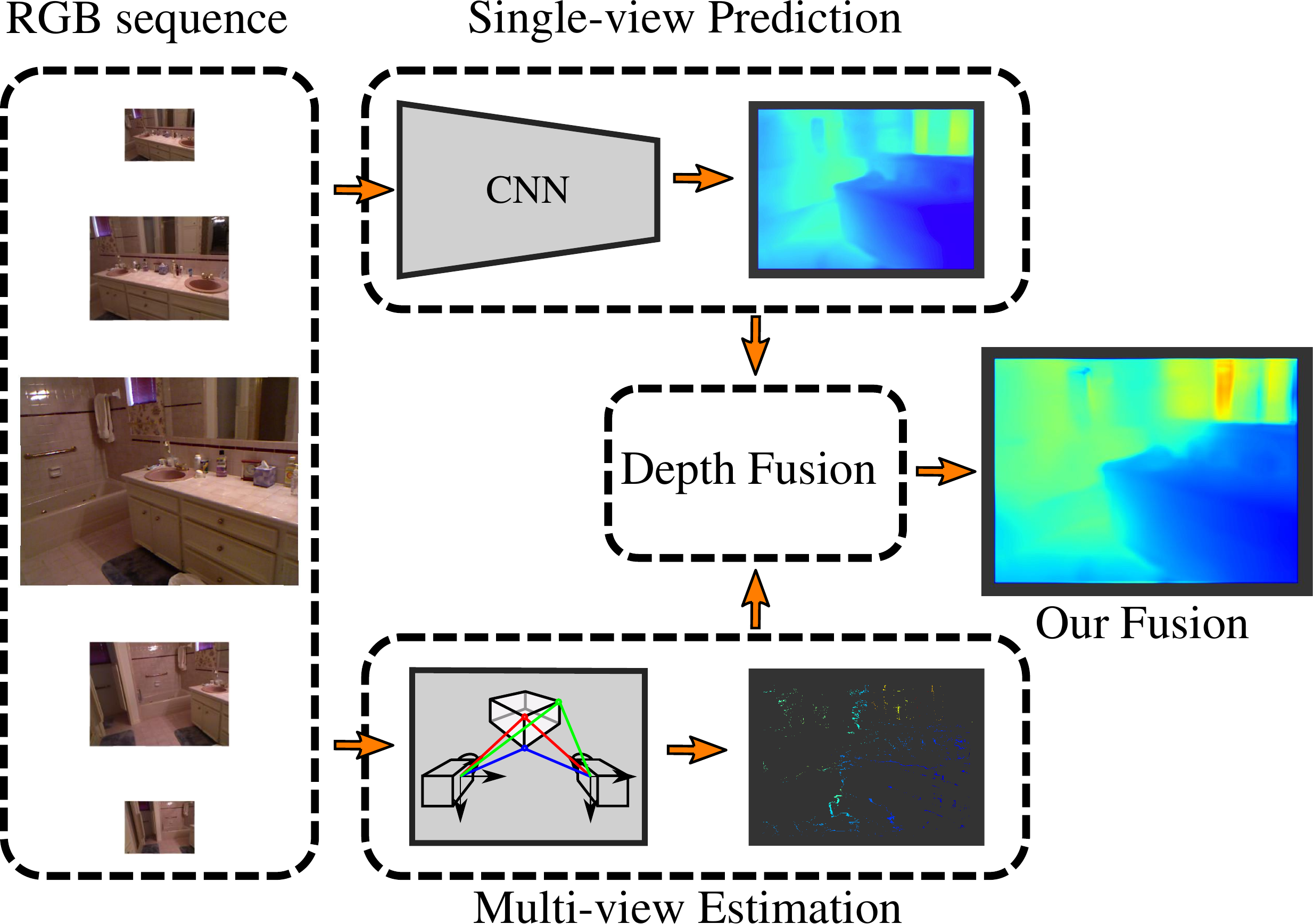}
\caption{\label{fig:overview} Overview of our proposal. The input is a set of overlapping monocular views. The learning-based single-view and geometry-based multi-view depth are fused, outperforming both of them. All the depth images are color-normalized for better comparison. This figure is best viewed in color.}
\end{figure}
The state of the art are the so-called direct mapping methods \cite{newcombe2011dtam}, that estimate an image depth by minimizing a regularized cost function based on the photometric error between corresponding pixels in several views. 
The accuracy of the multi-view depth estimation depends mainly on three factors: 1) The geometric configuration, with lower accuracies for low-parallax configurations; 2) the quality of the correspondences among views, that can only be reliably estimated for high-gradient pixels; and 3) the regularization function, typically the Total Variation norm, that is inaccurate for large texture-less areas. Due to this poor performance on large low-gradient areas, semi-dense maps are sometimes estimated only in high-gradient image pixels for visual direct SLAM (e.g., \cite{engel2014lsd}). Such semi-dense maps are accurate in high-parallax configurations but not a complete model of the viewed scene. Low-parallax configurations are mostly ignored in the visual SLAM literature.

An alternative method is single-view depth estimation, which has recently experienced  a qualitative improvement in its accuracy thanks to the use of deep convolutional networks \cite{eigen2015predicting}. Their accuracy is still lower than that of multi-view methods for high-texture and high-parallax points. But, as we will argue in this paper, they improve the accuracy of multi-view methods in low-texture areas due to the high-level feature extraction done by the deep networks --opposed to the low-level high-gradient pixels used by the multi-view methods. Interestingly, the errors in the estimated depth seem to be locally and not globally correlated since they come from the deep learning features. 

The main idea of this paper is to exploit the information of single and multi-view depth maps to obtain an improved depth even in low-parallax sequences and in low-gradient areas. Our contribution is an algorithm that fuses these complementary depth estimations. There are two main challenges in this task. First, the error distribution of the single-view estimation has several local modes, as it depends on the image content and not on the geometric configuration. Single and multi-view depth are hence related by a content-dependent 
deformation. 
Secondly, modeling the multi-view accuracy is not trivial when addressing general cases, including high and low-parallax configurations. 

We propose a method based on a weighted interpolation of the single-view local structure based on the quality and influence area of the multi-view semi-dense depth and evaluate its performance in two public datasets --NYU and TUM. The results show that our fusion algorithm improves over both individual single and multi-view approaches.

The rest of the paper is organized as follows. Section \ref{sec:related} describes the most relevant related work. Section \ref{sec:fusion} motivates and details the proposed algorithm for single and multi-view fusion. Section \ref{sec:experiments} presents our experimental results and, finally, Section \ref{sec:conclusions} contains the conclusions of this work.

\section{RELATED WORK}
\label{sec:related}

We classify the related work for dense depth estimation into two categories: methods based in multiple views of the scene and those which predict depth from one single image.

\subsection{Multi-View Depth}

In the multi-view depth estimation, \cite{newcombe2011dtam,graber2011online,stuhmer2010real} are the first works that achieved dense and real-time reconstructions from monocular sequences. Some of the most relevant aspects are the direct minimization of the photometric error --instead of the traditional geometric error of sparse reconstructions-- and the regularization of the multi-view estimation by adding the total variation (TV) norm to the cost function. 

TV regularization has low accuracy for large textureless areas, as shown recently in \cite{conchamanhattan,pinies2015dense,pinies2015too} among others. In order to overcome this \cite{conchamanhattan} proposes a piecewise-planar regularization; the plane parameters coming from multi-view superpixel triangulation \cite{concha2014using} or layout estimation \cite{hedau2009recovering}. \cite{pinies2015dense} proposes higher-order regularization terms that enforce piecewise affine constraints even in separated pixels. \cite{pinies2015too} selects the best regularization function among a set using sparse laser data. Building on \cite{conchamanhattan}, \cite{concha2015incorporating} adds the sparse data-driven 3D primitives of \cite{fouhey2013data} as a regularization prior. Compared to these works, our fusion is the first one where the information added to the multi-view depth is fully dense, data-driven and single-view; and hence it does not rely on additional sensors, parallax or Manhattan and piecewise-planar assumptions. It only relies on the network capabilities for the current domain, assuming that the test data follows the same distribution that the data used for training.

Due to the difficulty of estimating an accurate and fully dense map from monocular views there are several approaches that estimate only the depth for the highest-gradient pixels (e.g., \cite{engel2014lsd}). While this approach produces maps of higher density than the more traditional feature-based ones (e.g., \cite{mur2015orb}), they are still incomplete models of the scene and hence their applicability might be more limited.

\subsection{Single-View Depth}
Depth can be estimated from a single view using different image cues, for example focus (e.g., \cite{ens1993investigation}) or perspective (e.g., \cite{sturm1999method}). Learning-based approaches, as the one we use, basically discover RGB patterns that are relevant for accurate depth regression.

The pioneering work of Saxena {\em et~al.} \cite{saxena2009make3d} trained a MRF to model depth from a set of global and local image features. Before that, \cite{saxena2007depth} presented an early approach to depth prediction from monocular and stereo cues. Eigen {\em et~al.} \cite{eigen2014depth} presented a two deep convolutional neural network (CNN) stacked, one to predict global depth an the second one that refines it locally. Build upon this method, \cite{eigen2015predicting} recently presented a three scale convolutional network to estimate depth, surface normals and semantic labeling. Liu {\em et~al.} \cite{liu2015deep} use a unified continuous CRF-and-CNN framework to estimate depth. The CNN is used to learn the unary and pairwise potentials that the CRF uses for depth prediction. 

Based on \cite{eigen2015predicting}, \cite{li2016learning} incorporates mid-level features in its prediction using \textit{skip-layers}. It shows competitive results and a small batch-size training strategy that makes their network faster to train. \cite{chakrabarti2016depth} introduces a different method to predict depth from single-view using deep neural networks, showing that training the network with a much richer output improves the accuracy. \cite{cao2016estimating} formulates the depth prediction as a classification problem and the net output is a pixel-wise distribution over a discrete depth range. Finally, \cite{godard2016unsupervised} presents an unsupervised network for depth prediction using stereo images.

\section{SINGLE AND MULTI-VIEW DEPTH FUSION}
\label{sec:fusion}
State-of-the-art multi-view techniques have a strong dependency on high-parallax motion and heterogeneous-texture scenes. Only a reduced set of salient pixels that hold both constraints has a small error, and the error for the majority of the points is large and uncorrelated. In contrast,  single-view methods based on CNN networks achieve reasonable errors in all the image but they are locally correlated. 
Our proposal exploits the best properties of these two methods. Specifically, it  uses a deep convolutional network (CNN) to produce rough depth maps and fuses their structure with the results of a semi-dense multi-view depth method (Fig. \ref{fig:overview}).
\begin{table}

\centering
\begin{tabular}{c|c|c|}
\multicolumn{1}{c}{}\\
\cline{2-3}
   			& High-Gradient & Low-Gradient 	 \\ \hline
Multi-View 	& 	0.18	   &	1.02		 \\ \hline
Single-View & 	0.36	   & 	0.42         \\
\hline
\end{tabular}
\caption{ \label{tab:errdist} Median depth error [m] for single and multi-view depth estimation, and high and low-gradient pixels. This evaluation has been done in the sequence \textit{living\_room\_0030a} from the NYUv2 dataset (one of the sequences with higher parallax). The normalized threshold between high and low-gradient pixels is 0.35 (gray scale).}
\end{table}

Before delving into the technical aspects, we will motivate our proposal with some illustrative results. Table \ref{tab:errdist} shows the median depth error of the high-gradient and low-gradient pixels for a multi-view and single view reconstruction using a medium/high-parallax sequence of the NYUv2 dataset. For the multi-view reconstruction, the error for the low-gradient pixels increases by a factor of 2. Notice that the opposite happens for the single-view reconstruction: the error of high-gradient pixels is the one increasing by a factor of 2.  For this experiment, the  threshold used  to distinguish between high and low-gradient pixels is 0.35 in gray scale (where the maximum gradient would be 1).

\begin{figure}[t!]
\includegraphics[width=\linewidth]{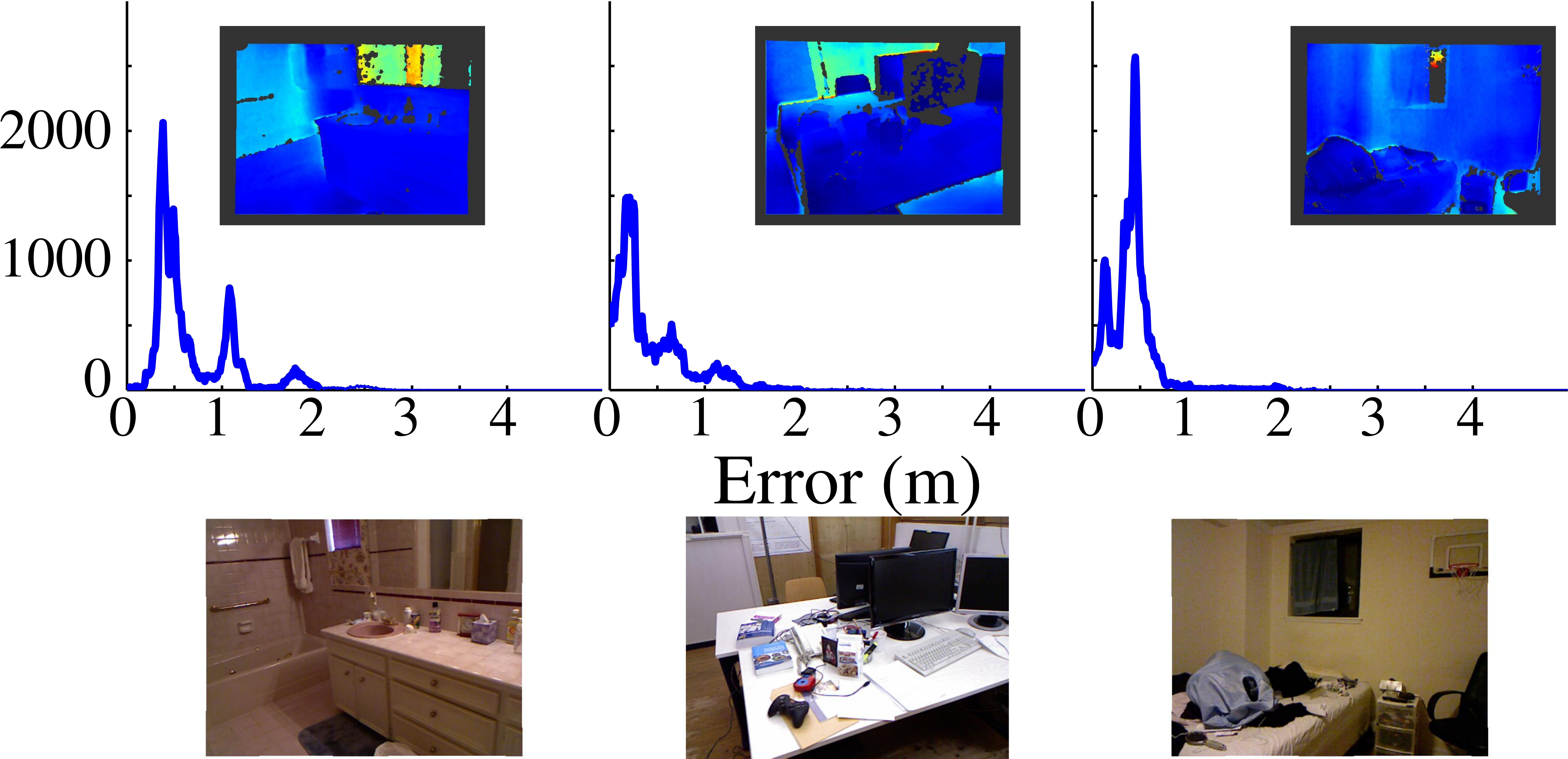}
\caption{\label{fig:hist}  Histogram of single-view depth error [m] for three sample sequences. Notice the multiple modes, each one corresponding to a local image structure, this can be seen in the error images in the top row of the figure.}
\end{figure}

Furthermore, the single-view depth error usually has a structure that indicates the presence of local correlations. For instance, Fig. \ref{fig:hist} shows the histogram of the single-view depth estimation error for three different sequences (two of the NYUv2 dataset and one of the TUM dataset). Notice that the error distribution is grouped in different modes, each one corresponding to an image segment. 

This effect is caused by the use of the high-level image features of the latest layers of the CNN network, that extend over dozens of pixels in the original image and hence over homogeneous texture areas. 
 The different nature of the errors can be exploited to outperform both individual estimations. This fusion, however, cannot be na\"ively implemented with a simple global model as it requires content-based deformations.

In the next subsections we detail the specific multi and single-view methods that we use in this work and our fusion algorithm.

\subsection{Multi-view Depth}
\label{sec:multi}
For the estimation of the multi-view depth we adopt a direct approach \cite{engel2014lsd}, that allows us to estimate a dense or semi-dense map in contrast to the more sparse maps of the feature-based approaches. In order to estimate the depth of a keyframe $\mathcal{I}_k$ we first select a set of $n$ overlapping frames $\{\mathcal{I}_1,\hdots,\mathcal{I}_o,\hdots,\mathcal{I}_n\}$ from the monocular sequence. After that, every pixel $x^k_l$ of the reference image $\mathcal{I}_k$ is first backprojected at an inverse depth ${\rho}$ and projected again in every overlapping image $\mathcal{I}_o$. 
\begin{equation} 
\label{eq:backandforthproj}
x^o_l = T_{ko}(x^k_l,\rho_l) = K R_{ko}^\top \left( \left(
\begin{array}{c}
\frac{K^{-1}x^k_l}{||K^{-1}x^k_l||}\\
\rho_l\\
\end{array}
\right) - t_{ko} \right),
\end{equation}
where $T_{ko}$,$R_{ko}$ and $t_{ko}$ are respectively the relative transformation, rotation and translation between the keyframe $\mathcal{I}_k$ and every overlapping frame $\mathcal{I}_o$. $K$ is the camera internal calibration matrix.

We define the total photometric error $C(\rho)$ as the summation of every photometric error $\epsilon_l$ between every pixel (or every high-gradient pixel if we want a semi-dense map) $x^k_l$ in the reference image $\mathcal{I}_k$ and its corresponding one $x^o_l$ in every other overlapping image $\mathcal{I}_o$ at an hypothesized inverse depth $\rho_l$, 
\begin{eqnarray} 
C(\rho) &=& \frac{1}{n} \sum_{o=1, o\neq k}^{n} \sum_{l=1}^{t} \epsilon_l(\mathcal{I}_k,\mathcal{I}_o,x^k_l,\rho_l).
\end{eqnarray}

The error $\epsilon_l(\mathcal{I}_k,\mathcal{I}_o,x^k_l,\rho_l)$ for each individual pixel $x^k_l$ is the difference between the photometric values of the pixel and its corresponding one
\begin{eqnarray} \label{eq:pherror}
\epsilon_l(\mathcal{I}_k,\mathcal{I}_o,x^k_l,\rho_l) &=& \mathcal{I}_k(x^k_l) - \mathcal{I}_o (x^o_l).
\end{eqnarray}

The estimated depth for every pixel $\hat{\rho} = (\hat{\rho_1} \ \hdots \ \hat{\rho_l} \ \hdots \  \hat{\rho_t} \ )^\top$ is obtained by the minimization of the total photometric error $C(\rho)$:
\begin{eqnarray} 
\hat{\rho} &=& \argmin_\rho C(\rho)
\end{eqnarray}

\subsection{Single-view Depth}
\label{sec:single}
For single-view depth estimation we use the Deep Convolutional Neural Network presented by Eigen {\em et~al.}, \cite{eigen2015predicting}. This network uses three stacked CNN to process the images in three different scales. The input to the network is the RGB keyframe $\mathcal{I}_k$. As we use the network structure and parameters released by the authors without further training, our input image size is $320\times240$. The output of the network is the predicted depth, that we will denote as $s$. The size of the output is $147\times109$, that we upsample in our pipeline in order to fuse it with the multi-view depth. 

The first scale CNN extract high-level features tuned for depth estimation. This CNN produces $64$ feature maps of size $19\times14$ that are the input, along with the RGB image, of the second scale CNN. This second stacked CNN refines the output of the first one with mid-level features to produce a first coarse depth map of size $74\times55$. This depth map is upsampled and feeds a third stacked CNN that does a local refinement of the depth. This final step is necessary, as the convolution and pooling steps of the previous layers filter out the high-frequency details. 

The first scale was initialized with two different pre-trained networks: the AlexNet \cite{krizhevsky2012imagenet} and the Oxford VGG \cite{simonyan2014very}. We use the VGG version, the most accurate one as reported by the authors. This network has been trained in indoor scenes with the NYUDepth v2 dataset \cite{silbermanECCV12}. As they used the official train/test splits of the dataset, so do we. We decided to use this neural network because it was the best-performing dense single-view method at the moment we started this work and still it is the one that keeps better trade off between quality and efficiency. We refer the reader to the original work \cite{eigen2015predicting} for more details on this part of our pipeline.

\subsection{Depth Fusion}

As we mentioned before, the objective is to fuse the output of each previous method while keeping the best properties of each of them: the single-view reliable local structure and the accurate, but semi-dense multi-view depth estimation.  
Let denote $s$ and $m$ to the single-view depth and the multi-view semi-dense depth estimation, respectively. $s$ is predicted as detailed in section \ref{sec:single} and $m = \frac{1}{\rho}$ is the inverse of the inverse depth estimated in section \ref{sec:multi}.

The fused depth estimation $f_{ij}$ for each pixel $(i,j)$ of a keyframe $\mathcal{I}_k$ is computed as a weighted interpolation of depths over the set of pixels in the multi-view depth image
\begin{equation}
\label{eq:fusion}
f_{ij} = \sum_{(u,v)\in \Omega} W^{m_{uv}}_{s_{ij}}(m_{uv}+(s_{ij}-s_{uv})),
\end{equation}
where $\Omega$ is the semi-dense set of pixels estimated by the multi-view algorithm (e.g. in a high-parallax sequence, they usually correspond with the high-gradient pixels).  The interpolation weights  $W^{m_{uv}}_{s_{ij}}$ model the likelihood for each pixel $(u,v)\in \Omega$  belonging to the same local structure as pixel $(i,j)$. The interpolation can be interpreted in two ways. First, the depth gradient $(s_{ij}-s_{uv})$ is added to each multi-view depth $m_{uv}$, i.e. we create depth map for each $m_{uv}$ with the structure of $s$ and then weigh them with pixel based weights. Second, for each depth $s_{ij}$ we modify it according to the weighted discrepancy between $(m_{uv}-s_{uv})$.

The key ingredient of this interpolation are the weights $W^{m_{uv}}_{s_{ij}}$ that model a 
deformation based on the local image structures. Each weight is computed as the product of four different factors. The first factor 
\begin{equation}
\tilde{W_1}^{m_{uv}}_{s_{ij}}=e^{\frac{-\sqrt{(i-u)^2+(j-v)^2))}}{\sigma_1}},
\end{equation} simply measures proximity based on the distance of the pixels $(i,j)$ and $(u,v)$. The  parameter  $\sigma_1$ controls the radius of proximity for each point. The remainder three factors depend on the structure of the single-view prediction $s$. The second factor 
\begin{equation}
\begin{split}
\tilde{W_2}^{m_{uv}}_{s_{ij}}= & \frac{1}{|\nabla_xs_{uv}-\nabla_xs_{ij}|+\sigma_2} \\ 
& \cdot  \frac{1}{|\nabla_ys_{uv}-\nabla_ys_{ij}|+\sigma_2}
\end{split}
\end{equation}measures the similarity of depth gradients and assigns larger weights to similar ones.  $\nabla_xs_{ij}$ and $\nabla_ys_{ij}$ represent the depth gradient in the $x$ and $y$ direction respectively at the pixel $(i,j)$.  $\sigma_2$ limits the influence of a point to avoid extremely high weights for very similar or identical gradients. We set it to 0.1 in the experiments. 

Finally, the factors $\tilde{W_3}^{m_{uv}}_{s_{ij}}$ and $\tilde{W_4}^{m_{uv}}_{s_{ij}}$ strengthen the influence between the points lying in the same plane and are defined as
\begin{equation}
\tilde{W_3}^{m_{uv}}_{s_{ij}}=e^{-|(s_{ij}+\nabla_xs_{ij}\cdot(u-i))))-s_{uv}|}+\sigma_3
\end{equation}
and
\begin{equation}
\tilde{W_4}^{m_{uv}}_{s_{ij}}=e^{-|(s_{ij}+\nabla_ys_{ij}\cdot(v-j))))-s_{uv}|}+\sigma_3,
\end{equation}where $\sigma_3$ sets a minimum weight to any point in $\Omega$. This is required to avoid vanishing weights when they are combined with $\tilde{W_1}^{m_{uv}}_{s_{ij}}$ and  $\tilde{W_2}^{m_{uv}}_{s_{ij}} $.  
\begin{figure}[t!]
\includegraphics[width=\linewidth]{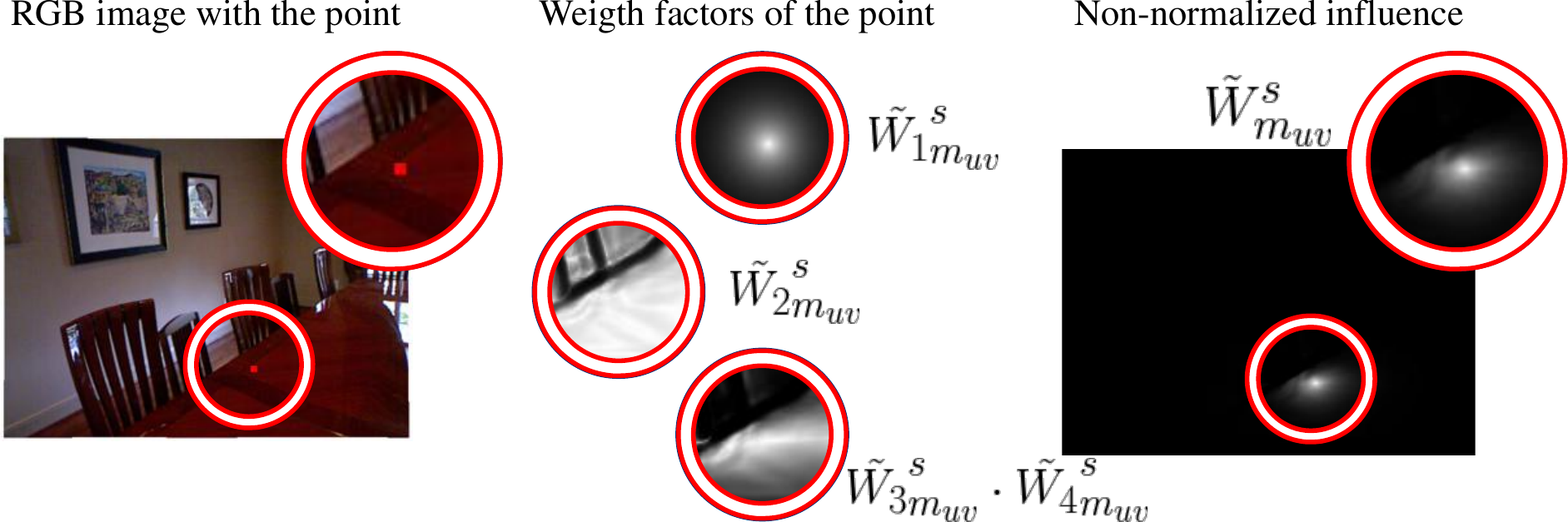}
\caption{\label{fig:singleinflu} Non-normalized influence of the highlighted red point in the image. \textit{First column:} RGB input image with a red point over the table, this point represent one pixel estimated by the multi-view algorithm. \textit{Second column:} each one of the weights calculated separately, the third and fourth weights are shown as a product for a more intuitive view.\textit{ Third column:} Non-normalized influence of the highlighted point in the RGB image. Notice how its influence is cut on the edge of the table. Figure best viewed in electronic format.}
\end{figure}

The product of this four factor makes a non-normalized weight for each pixel in $\Omega$
\begin{equation}
\tilde{W}_{m_{uv}}^{s_{ij}}=\prod_{n=1}^4 \tilde{W_n}_{m_{uv}}^{s_{ij}}
\end{equation}
and represents its area of influence. The parameters $\sigma_1$, $\sigma_2$ and $\sigma_3$ shape the area of influence and have to be selected to balance proximity, gradient and planarity and to avoid discontinuities in the result of the fusion. This was done empirically on a small set of three images. The values of the parameters are $15$, $0.1$ and $1e-3$, respectively, and we kept them fixed for all our experiments.

Fig. \ref{fig:singleinflu} shows this area for a point on an image and how it is computed. Notice how the influence expands around the point but is kept inside the same local structure (the table). Once all the factors has been computed, since all the pixels $(i,j)$ are influenced by all the pixels in $\Omega$ (see Eq. \ref{eq:fusion}), we normalize the weights for each single-view pixel so all the weights over a pixel $(i,j)$ sum 1.
\begin{equation}
W^{m_{uv}}_{s_{ij}}=\frac{\tilde{W}^{m_{uv}}_{s_{ij}}-\min_{(g,h)\in \Omega}{\tilde{W}^{m_{gh}}_{s_{ij}}}}{\sum_{(p,k)\in \Omega}{\tilde{W}^{m_{pk}}_{s_{ij}}-\min_{(g,h)\in \Omega}{\tilde{W}^{m_{gh}}_{s_{ij}}}}}
\end{equation}

\begin{figure}[t!]
\includegraphics[width=\linewidth]{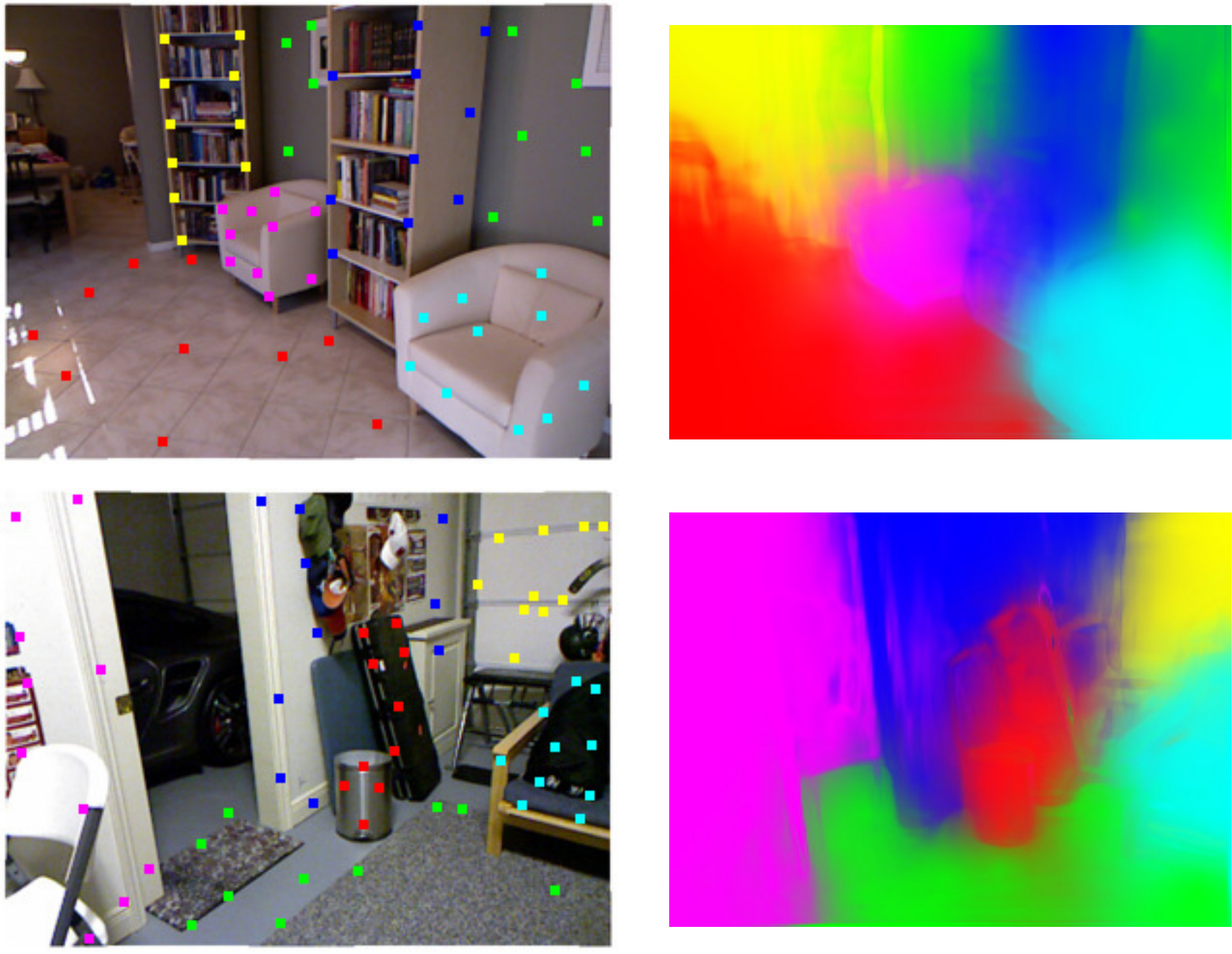}
\caption{\label{fig:globalinflu} Normalized influence area of the points. Notice how it expands around local structure areas given a set of points in $\Omega$. \textit{First column:} RGB image with the points of $\Omega$ labeled with different colors. \textit{Second column:} influence areas computed by our method.  Notice how this influence expands in areas with the same local structure but can be misled in areas where there is a lack of points or where the estimation from the neural net is not accurate enough. Figure best viewed in color.}
\end{figure}

\begin{figure}[t!]
\includegraphics[width=\linewidth]{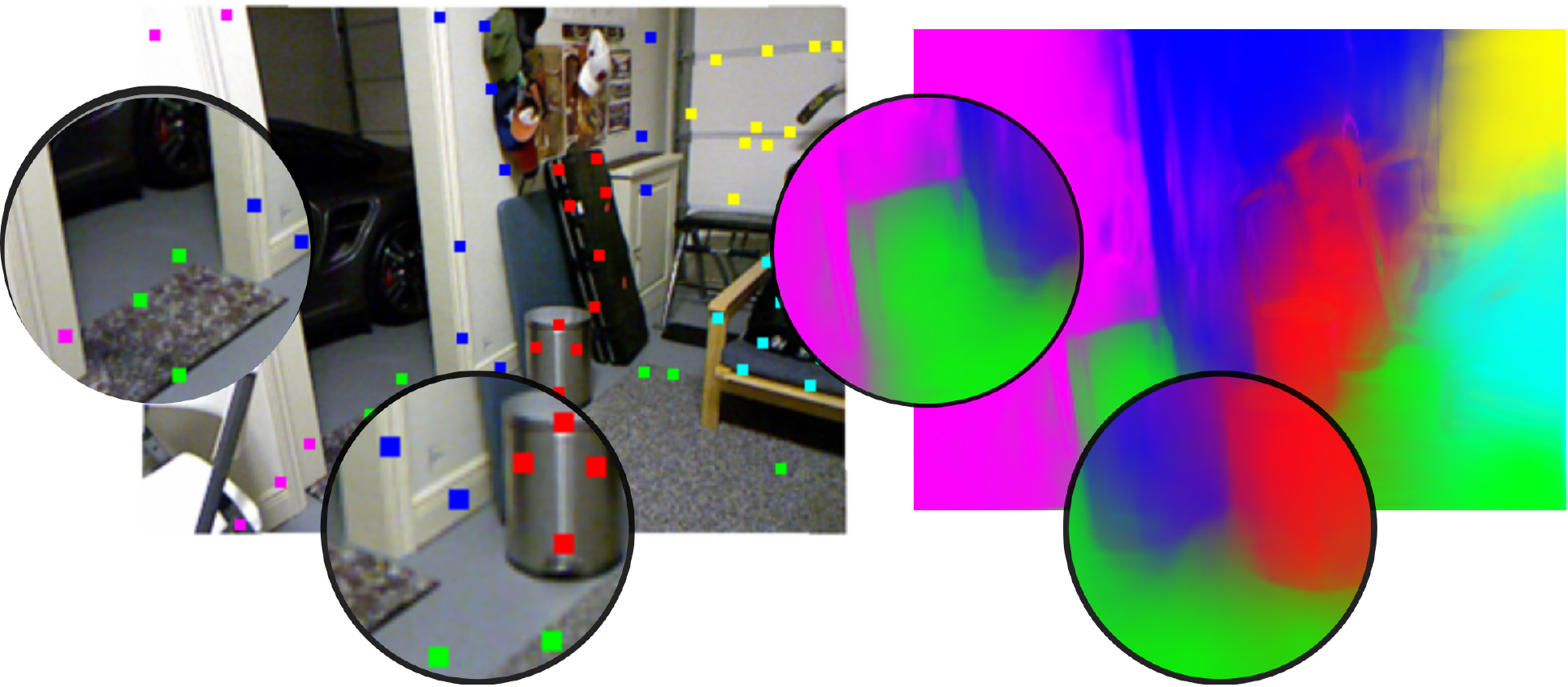}
\caption{\label{fig:globalinfludet} Detail of the influence area. Notice how it expands mainly in the areas with same local structure. Figure best viewed in color.}
\end{figure}

The normalized weights expand the local influence to the whole image (see Fig. \ref{fig:globalinflu} and Fig. \ref{fig:globalinfludet} for a more detailed view). Notice how the influence expands along planes even if the points in $\Omega$ do not reach the end of the plane; and is sharply reduced when the local structure changes. Once these influence weights have been calculated and normalized, the fusion depth estimation, $f$, for each point $(i,j)$ is a combination of all the selected points in $\Omega$, as presented in Eq. \ref{eq:fusion}. 

\subsection{Multi-view Low-Error Point Selection}

Up to now we have assumed that all the points in the multi-view semi-dense depth map $\Omega$ have low error. This is easily achievable in high-parallax sequences by using robust estimators --robust cost functions or RANSAC. However, it is problematic for the degenerate or quasi-degenerate low-parallax geometries that we also target in this paper. In this case, multi-view depths may contain large errors that will propagate to the fused depth map and it is necessary to filter them out. Unexpectedly, selecting high gradient pixels was not robust enough to remove points with large depth errors and we have developed a two step algorithm that takes into account photometric and geometric information in the first step and the single-view depth map in the second one. 

The first step selects a fixed percentage of the best correspondence candidates --the best $25\%$ in our experiments-- based on the product of a photometric and a geometric scores. On one hand, the photometric criterion focuses on the quality of the correspondences using image information. We  apply a modified version of the \emph{second best ratio}.We first extract the two closest matches for a pixel (smallest photometric errors according to Eq. \ref{eq:pherror}). We then compute the score as a function of the ratio between the distance of the two descriptors (a high ratio suggesting a good match) and the gradient of the distance function along the epipolar line (i.e., the error function presenting a distinct \emph{V-shape} around this match and suggesting spatial accuracy). On the other hand, the geometric score simply backpropagates the image correspondence error to the depth estimation, resulting in low scores for low-parallax correspondences.
 
In a second stage we also use the structure of the single-view reconstruction and apply RANSAC to estimate a spurious-free linear transformation between the multi and single-view points using only the points pre-filtered in the first stage. We apply this linear model along the entire image, consensus with outliers is found if small patches are used. This reduces further the number of spurious depth values from the multi-view algorithm. The result is a small set of low-error points that we use for the interpolation of the previous section. As mentioned before, in our experiments this algorithm behaves better than a geometric-only compatibility test, especially in the low-parallax sequences of the NYUv2 dataset.

\section{EXPERIMENTAL RESULTS}
\label{sec:experiments}

\begin{table*}
\scriptsize
\centering
\begin{tabular}{c|c|ccc|ccc|ccc|c|c|}
\multicolumn{1}{c}{}\\
\hhline{~~|---|---|---|~|-|}
\multicolumn{1}{c}{} & \multicolumn{1}{c}{} & \multicolumn{3}{|c|}{RMSE} & \multicolumn{3}{|c|}{SCALE INVARIANT} & \multicolumn{3}{|c|}{MEAN ERROR (m)}  & \quad & MEAN ERROR (m) \\
\hhline{~|-|-|-|-|-|-|-|-|-|-|~|-|}
      & Sequence   & TV & Eigen\cite{eigen2015predicting}  & Ours(auto)    & TV & Eigen\cite{eigen2015predicting}  & Ours(auto) & TV & Eigen\cite{eigen2015predicting}  & Ours(auto) &   \quad & Ours(man) \\
\hhline{-|-|-|-|-|-|-|-|-|-|-|~|-|}
 \parbox[t]{2mm}{\multirow{6}{*}{\rotatebox[origin=c]{90}{NYUDepth v2}}}
& \multicolumn{1}{r|}{bathroom\_0018}  & 1.458 & 0.852 & \textbf{0.793} & 0.405 & 0.150 &\textbf{0.145} & 1.174 & 0.692 & \textbf{0.612}
& \quad & 0.263\\
 & \multicolumn{1}{r|}{bedroom\_0013}  & 1.004& 0.550 & \textbf{0.482} & 0.212 & 0.139& \textbf{0.136}& 0.690 & 0.441  & \textbf{0.344}
 & \quad & 0.163\\
 & \multicolumn{1}{r|}{dining\_room\_0032}  & 2.212 & 0.710 & \textbf{0.694} & 0.416& 0.209&\textbf{0.204} &  1.797 & 0.581 & \textbf{0.554}
 & \quad & 0.318\\
 & \multicolumn{1}{r|}{kitchen\_0032}  & 3.599 & 1.621 & \textbf{1.572} & 0.812& 0.592& \textbf{0.583}&  2.920 & 1.222 & \textbf{1.183}
 & \quad & 0.805\\
 & \multicolumn{1}{r|}{living\_room\_0025} & 1.073 & 0.620 & \textbf{0.597} & 0.289 & 0.236& \textbf{0.219}&  0.798 & 0.471 & \textbf{0.435}
& \quad & 0.289\\
 & \multicolumn{1}{r|}{living\_room\_0030a} & 1.031 & 0.818 & \textbf{0.792} & 0.411& 0.228 & \textbf{0.219}&  0.849 & 0.532 & \textbf{0.440}
 & \quad & 0.329\\
\hhline{==|===|===|===|~|=|}
 \parbox[t]{2mm}{\multirow{2}{*}{\rotatebox[origin=c]{90}{TUM}}} & \multicolumn{1}{r|}{fr1\_desk} & 1.581 & 0.433 & \textbf{0.410} &  0.255 & 0.121 & \textbf{0.103} &  1.211 & 0.317 & \textbf{0.294}
 & \quad & 0.154\\
 & \multicolumn{1}{r|}{fr1\_room}  &  1.467 & 0.323 & \textbf{0.301} & 0.167 & 0.092 & \textbf{0.081} &  1.163 & 0.231 & \textbf{0.207}
  & \quad & 0.102 \\
\hhline{-|-|-|-|-|-|-|-|-|-|-|~|-|}
\end{tabular}

\caption{\label{tab:results_all} \textit{Left table: } Error metrics for the NYUv2 and TUM datasets. For each sequence and metric we compare the TV-regularized multi-view depth, the single-view depth \cite{eigen2015predicting} and our fused depth.  \textit{Right table: }Mean error for the fused depth with manual multi-view point selection. (The evaluation has been performed in the first 100 frames of each sequence)}
\end{table*}

\begin{figure*}[t!]
\centering
\includegraphics[width=0.8\linewidth]{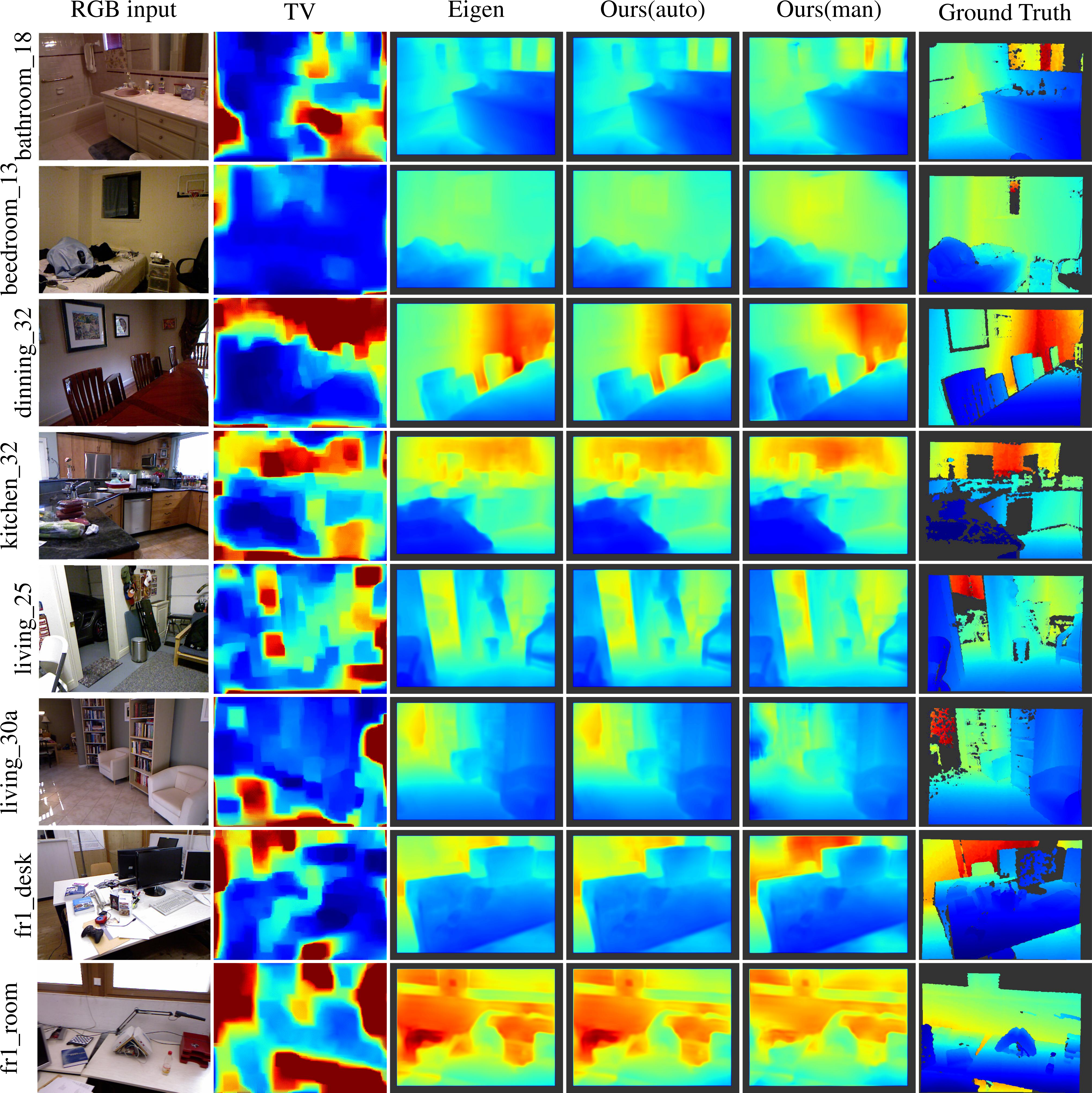}
\caption{\label{fig:results} The first six rows are depth images for the NYUDepth v2 dataset \cite{silbermanECCV12} and the last two rows are for the TUM Dataset \cite{sturm12iros}. Color ranges are row-normalized to facilitate the comparison between different methods. \textit{First column} RGB keyframe, \textit{second column} TV-regularized multi-view depth, \textit{third column} single-view depth, \textit{fourth column} our depth fusion with automatic multi-view point selection, \textit{fifth column} our depth fusion with manual multi-view point selection, and \textit{sixth column} ground truth. Figure best viewed in electronic format. }
\end{figure*}
In this section we evaluate the algorithm and compare its performance against two state-of-the-art methods: multi-view direct mapping using TV regularization (implemented following \cite{newcombe2011dtam,handa2011applications}) and the single-view depth estimation using the network of \cite{eigen2015predicting}. We have selected two datasets with different properties. The first one is the NYUv2 Depth Dataset \cite{silbermanECCV12}, a general dataset aimed at image segmentation evaluation and hence likely to contain low-parallax and low-texture sequences. We analyze results in  six sequences from the test set (i.e. the single-view net had not been trained on these sequences) selected just to include different types of rooms. The second one is the TUM RGB-D SLAM Dataset \cite{sturm12iros}, a dataset oriented to visual SLAM and then likely to present a bias benefiting multi-view depth. In this case, we evaluated two sequences selected randomly.

We run our algorithm in a $320\times240$ subsampled version of the images, as this is the size of the single-view neural network given by the authors. We also run our multi-view depth estimation at this image size, and upsample the fused depth to $640\times480$ in order to compare it against the ground truth D channel from the kinect camera.

As our aim is to evaluate 
the accuracy of the depth estimation, we will assume that camera poses are known for the multi-view estimation. In the TUM RGB-D SLAM Dataset \cite{sturm12iros} we use the ground truth camera poses.
In the NYUv2 Depth Dataset sequences we estimate them using the RGB-D Dense Visual Odometry by Gutierrez-Gomez {\em et~al.} \cite{gutierrezinverse}. These camera poses will remain fixed and used to create the multi-view depth maps. As mentioned before, the parameters of the fusion algorithm were experimentally set prior to the evaluation on a small separate set of images.

To evaluate the methods, we computed three different metrics, the RMSE, the Mean Absolute Error in meters and the scale invariant error proposed in \cite{eigen2014depth} \mbox{$\frac{1}{n}\sum_{i}d_i^2 - \frac{1}{n^2}(\sum_{i}d_i)^2$} where $d$ is $(\log(y)-\log(y^*))$, $y$ and $y^*$ are the ground truth depth and the estimated depth respectively. The results  are summarized in Table \ref{tab:results_all}. Our method outperforms the TV regularization in both datasets obtaining an average improvement over 50\% with respect to the mean of the error in meters. As expected, the TV regularization performs better in the TUM sequences and achieves lower errors, but in terms of improvement there seems not to be big differences between both datasets.
Our fusion of depths also outperforms the single-view depth reconstruction, the improvement being $10\%$ on average. Both methods perform similarly in both datasets, but except in one sequence, our method is always better or as good as the deep single-view reconstruction.  Notice that the improvement does not come exclusively from scale correction; the scale invariant error shows that our method improves the structure estimation in both the single and multi-view cases. %

The right-most colum of Table  \ref{tab:results_all} shows the depth errors when the set of multi-view points does not contain outliers. We selected them using the ground-truth data from the D channel, and keeping only those points whose depth error was lower than $10$cm. The results are for all sequences better than any method attaining improvements around $70\%$ and $38\%$  with respect to TV and \cite{eigen2015predicting}, respectively. Although expected, this result highlights the impact of multi-view outliers and the need for good point selection. It also provides an upper bound and shows that there is still room for improvement in this latest part of our algorithm.  In Table \ref{tab:weight_comp} we show an experiment to better understand the contribution of each weight of our algorithm. For this evaluation we have considered the spurious-free set of multi-view points in order to avoid the influence of noise. It can be seen that using all the weights has an average of 9.8\% improvement in mean absolute error with respect to using just $W_1$ and a 6.5\% of improvement with respect to using $W_1$ and $W_2$.

\begin{table}
    \setlength{\tabcolsep}{5pt} %

 \scriptsize
 \centering
 \begin{tabular}{r|ccc|ccc|}
 \multicolumn{1}{c}{}\\
 \hhline{~|---|---|}
 \multicolumn{1}{r|}{} &  \multicolumn{3}{c|}{SCALE INVARIANT} & \multicolumn{3}{c|}{MEAN ERROR (m)}  \\
 \hhline{~|-|-|-|-|-|-|}
   & $W_1$ & $W_1 \cdot W_2$  & $\prod_{i=1}^{4}W_i$ & $W_1$ & $W_1 \cdot W_2$  & $\prod_{i=1}^{4}W_i$ \\
 \hhline{~|-|-|-|-|-|-|}
  NYUv2 & 0.224 & 0.216 & \textbf{0.208} &  0.390 & 0.376 & \textbf{0.353}\\
 \hhline{-|-|-|-|-|-|-|}
  TUM & 0.098 & 0.088 & \textbf{0.064} &  0.145 & 0.142 & \textbf{0.128}\\
 \hhline{~|-|-|-|-|-|-|}
 \end{tabular}
 \caption{\label{tab:weight_comp} Mean of error metrics for the NYUv2 and TUM datasets. For each sequence and metric we compare the fusion with the only use of the weight $W_1$, the use of $W_1\cdot W_2$. and all the weights together.}
  \end{table}
\begin{figure}[t!]
\centering
\includegraphics[width=0.95\linewidth]{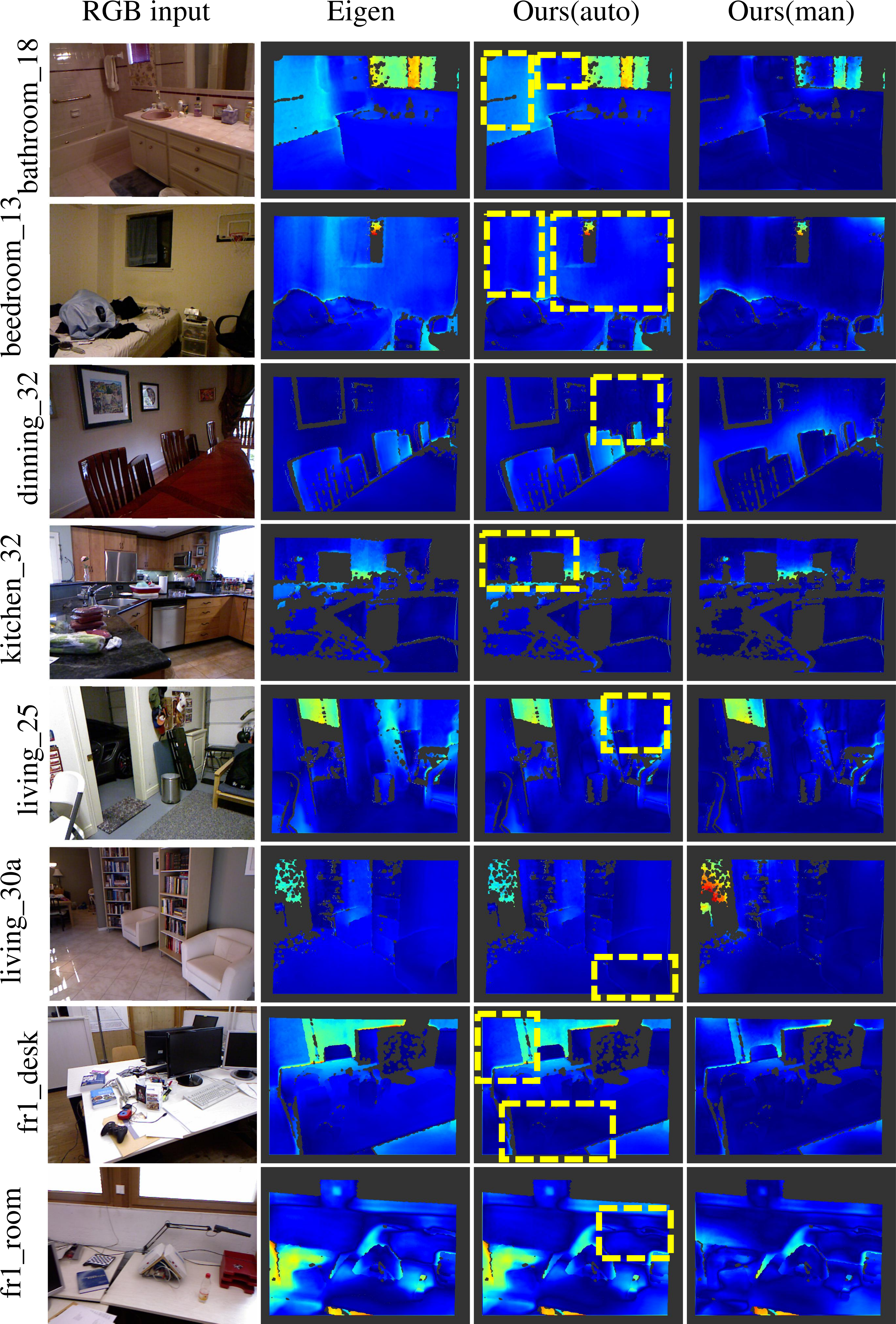}
\caption{\label{fig:erros} The first six rows are error images (predicted depth - ground truth) for the NYUDepth v2 dataset \cite{silbermanECCV12} and the last two rows are for the TUM Dataset \cite{sturm12iros}. Color ranges are row-normalized to facilitate the comparison between different methods. Darker blue is better. \textit{First column} RGB keyframe, \textit{second column} single-view depth, \textit{third column} our depth fusion with automatic multi-view point selection, \textit{fourth column} our depth fusion with manual multi-view point selection. In the third column, in yellow, are highlighted the areas where the improvement of our method can be easily appreciated with respect to single-view's error. Figure best viewed in electronic format.
}
\end{figure}
\begin{figure*}[t]
\centering
\includegraphics[width=0.9\linewidth]{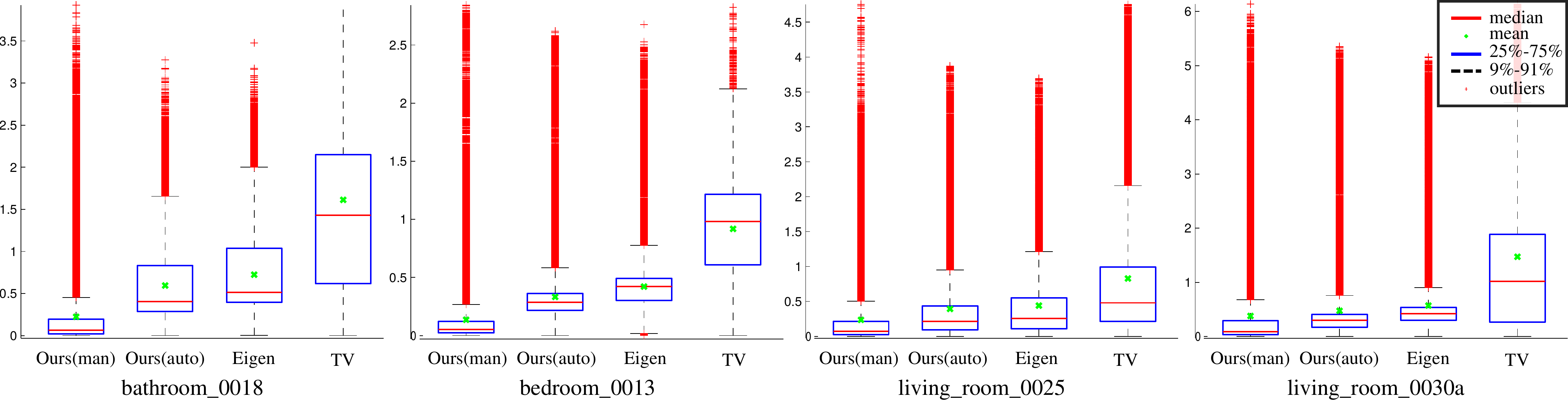}
\caption{\label{fig:boxplot} Box-and-Whiskers plots of the pixel error distribution for four of our test scenes. From left to right: Our method with manual point selection, our method with automatic point selection, single-view depth from Eigen {\em et~al.} \cite{eigen2015predicting} and TV-regularized multi-view depth.}
\label{fig:boxplots}
\end{figure*}
Finally, we present the results of some randomly picked images for each sequence of each dataset. Fig. \ref{fig:results}  shows the obtained depth images for the NYUDepth v2 and the TUM datasets. The improvement with respect to the regularized multi-view approach is clear visually since the depth structure is much more consistent.  Improvements with respect to single-view images are more subtle and are best viewed by looking at the corresponding depth error images of Fig. \ref{fig:erros}. Usually, the improvement comes from a better relative placement of some local structure. For instance, the walls are darker in the error images (see the bathroom\_18, bedroom\_13 or fr1\_desk in Fig. \ref{fig:erros}). The effect is more evident when the multi-view points were selected based on the ground truth. This better alignment of local structures reduces the error, as can be seen in the per-sequence error boxplots of Fig. \ref{fig:boxplots}.

\section{CONCLUSIONS}
\label{sec:conclusions}
In this paper we have presented an algorithm for dense depth estimation by fusing 1) the multi-view depth estimation from a direct mapping method, and 2) the single-view depth that comes from a deep convolutional network trained on RGB-D images. Our approach selects a set of the most accurate points from the multi-view reconstruction and fuses them with the dense single-view estimation. It is worth remarking that the single-view depth errors do not depend on the geometric configuration but on the image content and hence the transformation is not geometrically rigid and varies locally. The estimation of this alignment is our main contribution and the most challenging aspect of this research.

Our experiments show that our proposal improves over the state of the art (Eigen et al. \cite{eigen2015predicting} for single-view depth and direct mapping plus TV regularization for multi-view depth). Contrary to other approaches, the single-view depth we use is entirely data-driven and hence does not rely on any scene assumption. As mentioned, we take the network of \cite{eigen2015predicting} as our single-view baseline, because of its availability and its excellent accuracy-cost ratio. However, our fusion algorithm is independent of the specific network and could be used with any of the single-view approaches mentioned in Section \ref{sec:related}. Future work will, as suggested by the results, try to improve the multi-view points selection and the fusion of both images using, for instance, iterative procedures  or segmentation-based fusion.

\bibliographystyle{ieeetr}
\bibliography{chema.bib}

\begin{thebibliography}{10}

\bibitem{newcombe2011dtam}
R.~A. Newcombe, S.~J. Lovegrove, and A.~J. Davison, ``{DTAM: Dense tracking and
  mapping in real-time},'' in {\em Computer Vision (ICCV), 2011 IEEE
  International Conference on}, pp.~2320--2327, IEEE, 2011.

\bibitem{engel2014lsd}
J.~Engel, T.~Sch{\"o}ps, and D.~Cremers, ``{LSD-SLAM: Large-scale direct
  monocular SLAM},'' in {\em European Conference on Computer Vision},
  pp.~834--849, Springer, 2014.

\bibitem{eigen2015predicting}
D.~Eigen and R.~Fergus, ``Predicting depth, surface normals and semantic labels
  with a common multi-scale convolutional architecture,'' in {\em Proceedings
  of the IEEE International Conference on Computer Vision}, pp.~2650--2658,
  2015.

\bibitem{graber2011online}
G.~Graber, T.~Pock, and H.~Bischof, ``Online {3D} reconstruction using convex
  optimization,'' in {\em 2011 IEEE International Conference on Computer Vision
  Workshops}, pp.~708--711, IEEE, 2011.

\bibitem{stuhmer2010real}
J.~St{\"u}hmer, S.~Gumhold, and D.~Cremers, ``Real-time dense geometry from a
  handheld camera,'' in {\em Joint Pattern Recognition Symposium}, pp.~11--20,
  Springer, 2010.

\bibitem{conchamanhattan}
A.~Concha, W.~Hussain, L.~Montano, and J.~Civera, ``{Manhattan and
  Piecewise-Planar Constraints for Dense Monocular Mapping},'' in {\em
  Robotics: Science and Systems}, 2014.

\bibitem{pinies2015dense}
P.~Pinies, L.~M. Paz, and P.~Newman, ``{Dense mono reconstruction: Living with
  the pain of the plain plane},'' in {\em 2015 IEEE International Conference on
  Robotics and Automation}, pp.~5226--5231, 2015.

\bibitem{pinies2015too}
P.~Pini{\'e}s, L.~M. Paz, and P.~Newman, ``{Too much TV is bad: Dense
  reconstruction from sparse laser with non-convex regularisation},'' in {\em
  2015 IEEE International Conference on Robotics and Automation (ICRA)},
  pp.~135--142, IEEE, 2015.

\bibitem{concha2014using}
A.~Concha and J.~Civera, ``{Using superpixels in monocular SLAM},'' in {\em
  Robotics and Automation (ICRA), 2014 IEEE International Conference on},
  pp.~365--372, IEEE, 2014.

\bibitem{hedau2009recovering}
V.~Hedau, D.~Hoiem, and D.~Forsyth, ``Recovering the spatial layout of
  cluttered rooms,'' in {\em 2009 IEEE 12th international conference on
  computer vision}, pp.~1849--1856, IEEE, 2009.

\bibitem{concha2015incorporating}
A.~Concha, W.~Hussain, L.~Montano, and J.~Civera, ``Incorporating scene priors
  to dense monocular mapping,'' {\em Autonomous Robots}, vol.~39, no.~3,
  pp.~279--292, 2015.

\bibitem{fouhey2013data}
D.~F. Fouhey, A.~Gupta, and M.~Hebert, ``Data-driven 3d primitives for single
  image understanding,'' in {\em Proceedings of the IEEE International
  Conference on Computer Vision}, pp.~3392--3399, 2013.

\bibitem{mur2015orb}
R.~Mur-Artal, J.~Montiel, and J.~D. Tardos, ``{ORB-SLAM: a versatile and
  accurate monocular SLAM system},'' {\em Robotics, IEEE Transactions on},
  vol.~31, no.~5, pp.~1147--1163, 2015.

\bibitem{ens1993investigation}
J.~Ens and P.~Lawrence, ``An investigation of methods for determining depth
  from focus,'' {\em IEEE Transactions on pattern analysis and machine
  intelligence}, vol.~15, no.~2, pp.~97--108, 1993.

\bibitem{sturm1999method}
P.~Sturm and S.~Maybank, ``A method for interactive 3d reconstruction of
  piecewise planar objects from single images,'' in {\em The 10th British
  machine vision conference (BMVC'99)}, pp.~265--274, 1999.

\bibitem{saxena2009make3d}
A.~Saxena, M.~Sun, and A.~Y. Ng, ``{Make3D: Learning 3D scene structure from a
  single still image},'' {\em IEEE transactions on pattern analysis and machine
  intelligence}, vol.~31, no.~5, pp.~824--840, 2009.

\bibitem{saxena2007depth}
A.~Saxena, J.~Schulte, and A.~Y. Ng, ``Depth estimation using monocular and
  stereo cues.,'' in {\em IJCAI}, vol.~7, 2007.

\bibitem{eigen2014depth}
D.~Eigen, C.~Puhrsch, and R.~Fergus, ``Depth map prediction from a single image
  using a multi-scale deep network,'' in {\em Advances in Neural Information
  Processing Systems}, pp.~2366--2374, 2014.

\bibitem{liu2015deep}
F.~Liu, C.~Shen, and G.~Lin, ``Deep convolutional neural fields for depth
  estimation from a single image,'' in {\em IEEE Conference on Computer Vision
  and Pattern Recognition}, pp.~5162--5170, 2015.

\bibitem{li2016learning}
J.~Li, R.~Klein, and A.~Yao, ``Learning fine-scaled depth maps from single rgb
  images,'' {\em arXiv preprint arXiv:1607.00730}, 2016.

\bibitem{chakrabarti2016depth}
A.~Chakrabarti, J.~Shao, and G.~Shakhnarovich, ``Depth from a single image by
  harmonizing overcomplete local network predictions,'' in {\em Advances in
  Neural Information Processing Systems}, pp.~2658--2666, 2016.

\bibitem{cao2016estimating}
Y.~Cao, Z.~Wu, and C.~Shen, ``Estimating depth from monocular images as
  classification using deep fully convolutional residual networks,'' {\em arXiv
  preprint arXiv:1605.02305}, 2016.

\bibitem{godard2016unsupervised}
C.~Godard, O.~Mac~Aodha, and G.~J. Brostow, ``Unsupervised monocular depth
  estimation with left-right consistency,'' {\em arXiv preprint
  arXiv:1609.03677}, 2016.

\bibitem{krizhevsky2012imagenet}
A.~Krizhevsky, I.~Sutskever, and G.~E. Hinton, ``Imagenet classification with
  deep convolutional neural networks,'' in {\em Advances in neural information
  processing systems}, pp.~1097--1105, 2012.

\bibitem{simonyan2014very}
K.~Simonyan and A.~Zisserman, ``Very deep convolutional networks for
  large-scale image recognition,'' {\em arXiv preprint arXiv:1409.1556}, 2014.

\bibitem{silbermanECCV12}
P.~K. Nathan~Silberman, Derek~Hoiem and R.~Fergus, ``Indoor segmentation and
  support inference from {RGBD Images},'' in {\em ECCV}, 2012.

\bibitem{sturm12iros}
J.~Sturm, N.~Engelhard, F.~Endres, W.~Burgard, and D.~Cremers, ``{A benchmark
  for the evaluation of RGB-D SLAM systems},'' in {\em Intelligent Robots and
  Systems (IROS), 2012 IEEE/RSJ International Conference on}, pp.~573--580,
  IEEE, 2012.

\bibitem{handa2011applications}
A.~Handa, R.~A. Newcombe, A.~Angeli, and A.~J. Davison, ``Applications of
  legendre-fenchel transformation to computer vision problems,'' {\em
  Department of Computing at Imperial College London. DTR11-7}, vol.~45, 2011.

\bibitem{gutierrezinverse}
D.~Guti{\'e}rrez-G{\'o}mez, W.~Mayol-Cuevas, and J.~Guerrero, ``{Inverse Depth
  for Accurate Photometric and Geometric Error Minimisation in RGB-D Dense
  Visual Odometry},'' in {\em Robotics and Automation (ICRA), 2015 IEEE
  International Conference on}, pp.~83--89, IEEE, 2015.

\end{thebibliography}

\end{document}